\pdfoutput=1
\documentclass[11pt]{article}

\usepackage{naacl2021}

\usepackage{times}
\usepackage{latexsym}

\usepackage[T1]{fontenc}
\usepackage[utf8]{inputenc}

\usepackage{microtype}

\usepackage{graphicx}
\usepackage{soul}
\usepackage{booktabs}

\title{Translational NLP: A New Paradigm and General Principles \\for Natural Language Processing Research}

\author{Denis Newman-Griffis$^1$ \ \ \ \ \ Jill Fain Lehman$^2$\ \ \ \ \ \ Carolyn Ros\'{e}$^3$ \ \ \ \ \ Harry Hochheiser$^1$\\
$^1$Department of Biomedical Informatics, University of Pittsburgh, USA\\
$^2$Human-Computer Interaction Institute, Carnegie Mellon University, USA\\
$^3$Language Technologies Institute, Carnegie Mellon University, USA\\
{\small \tt \{dnewmangriffis, harryh\}@pitt.edu, \{jfl, cprose\}@cs.cmu.edu}
}

\begin{document}
\maketitle
\begin{abstract}
Natural language processing (NLP) research combines the study of universal principles, through basic science, with applied science targeting specific use cases and settings. However, the process of exchange between basic NLP and applications is often assumed to emerge naturally, resulting in many innovations going unapplied and many important questions left unstudied. We describe a new paradigm of {\it Translational NLP}, which aims to structure and facilitate the processes by which basic and applied NLP research inform one another. Translational NLP thus presents a third research paradigm, focused on understanding the challenges posed by application needs and how these challenges can drive innovation in basic science and technology design. We show that many significant advances in NLP research have emerged from the intersection of basic principles with application needs, and present a conceptual framework outlining the stakeholders and key questions in translational research. Our framework provides a roadmap for developing Translational NLP as a dedicated research area, and identifies general translational principles to facilitate exchange between basic and applied research.
\end{abstract}

\section{Introduction}

Natural language processing (NLP)  lies at the intersection of basic
science and applied technologies. However, translating innovations in basic NLP
methods to successful applications remains a difficult
task in which failure points often appear late in the development process,
delaying or preventing potential impact in research and industry.  Application
challenges range widely, from changes in data distributions \cite{Elsahar2019} to
computational bottlenecks \cite{Desai2020} and integration with domain expertise
\cite{Rahman2020}.  When unanticipated, such challenges can be fatal to
applications of new NLP methodologies, leaving exciting innovations with minimal
practical impact. Meanwhile, real-world applications may rely on regular
expressions \cite{Anzaldi2017} or unigram frequencies \cite{Slater2017} when more
sophisticated methods would yield deeper insight.
When successful translations of basic NLP insights into practical applied technologies do occur, the factors contributing to this success are rarely analyzed, limiting our ability to learn how to enable the next project and the next technology.

\begin{figure}
    \centering
    \includegraphics[width=0.48\textwidth]{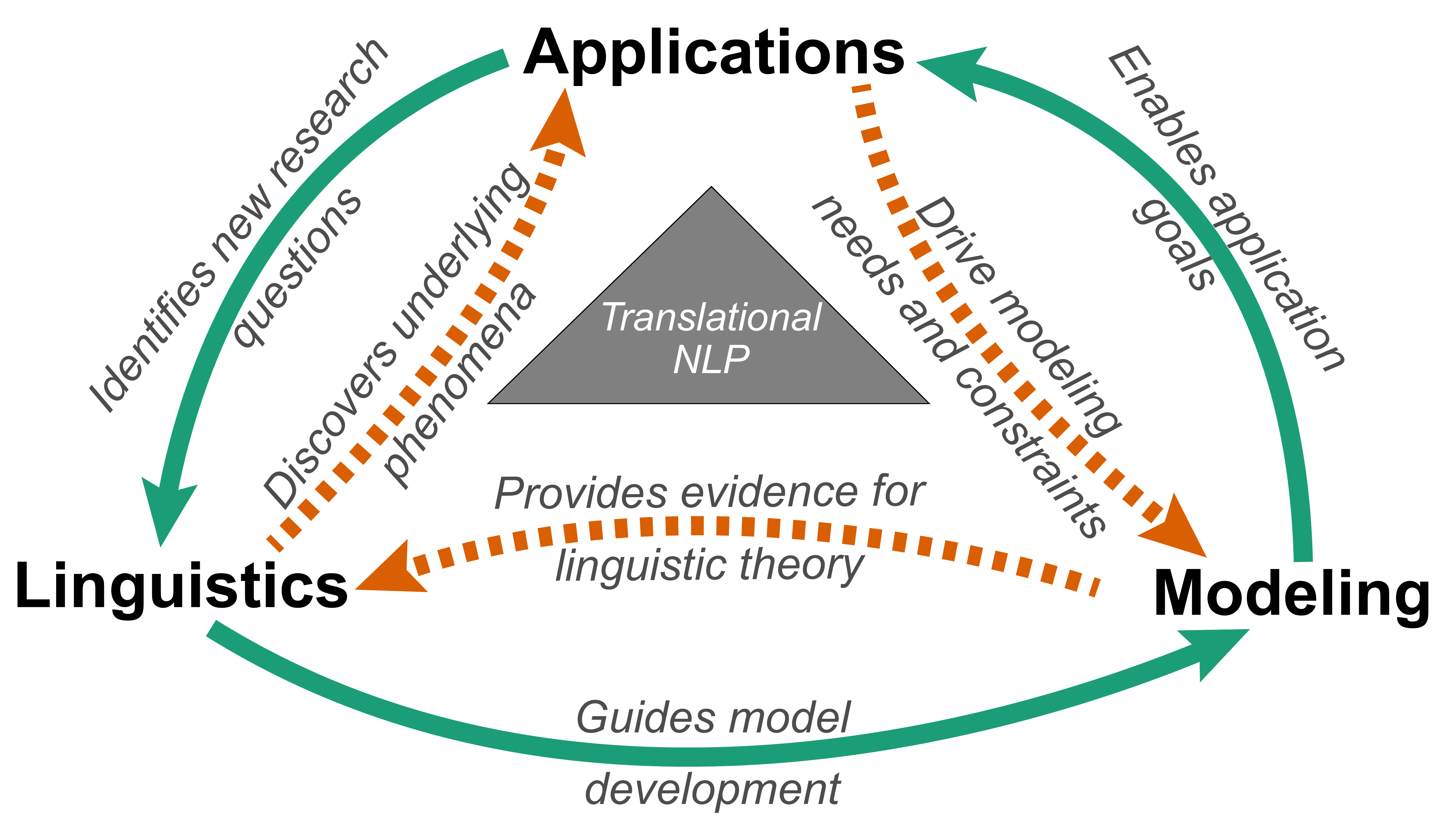}
    \caption{Interactions between linguistic theory, model development, and
    applications in NLP research. Solid lines indicate moving from basic research to applications, and dashed lines indicate how applied research feeds back into basic study. Translational NLP develops processes to realize this exchange.}
    \label{fig:research-network}
\end{figure}

We argue for a third kind of NLP research, which we call {\it Translational NLP}. Translational NLP research aims to understand why one translation succeeds while another fails, and to develop general, reusable processes to facilitate more (and easier) translation between basic NLP advances and real-world application settings. Much NLP research already includes translational insights, but often considers them properties of a specific application rather than generalizable findings that can advance the field.
This paper illustrates why \textit{general principles} of the translational process
enhance mutual exchange between linguistic inquiry, model development, and application research
(illustrated in Figure~\ref{fig:research-network}),
and are key drivers of NLP advances.

We present a conceptual framework for Translational NLP, with
specific elements of the translational process that are key to successful applications, each of which presents distinct areas for research.
Our framework provides a concrete path for designing use-inspired
basic research so that research products can effectively be turned into
practical technologies, and provides the tools to understand why a technology translation succeeds or fails. A translational perspective further enables factorizing ``grand challenge'' research
questions into clearly-defined pieces, producing intermediate results and
driving new basic research questions.
Our paper makes the following contributions:
\begin{itemize}
\item We characterize the stakeholders involved in the process of translating
      basic NLP advances to applications, and identify the roles they play in
      identifying new research problems (\S\ref{sec:stakeholders}).
\item We present a general-purpose checklist to use as a starting point for the  translational
      process, to help integrate basic NLP innovations into applications and
      to identify basic research opportunities arising from application needs
      (\S\ref{sec:checklist}).
\item We present a case study in the medical domain illustrating how
      the elements of our Translational NLP framework can lead to new challenges for
      basic, applied, and translational NLP research (\S\ref{sec:case-studies}).
\end{itemize}

\section{Defining Translational NLP}

\subsection{A third type of research}

A long history of distinguishing between basic and applied
research \cite{Bush1945,Shneiderman2016} has noted that these terms are often
relative; one researcher's
basic study is the application of another's theory. In practice, basic and
applied research in NLP are endpoints of a spectrum, rather than discrete
categories.  As \textit{use-inspired} research, most NLP studies incorporate
elements of both basic and applied research. We therefore define our key
terms for this paper as follows:

\textbf{Basic research} Basic NLP research is focused on universal principles:
linguistically-motivated study that guides model design (e.g.,
\citet{Recasens2009} for coreference, \citet{Koulompis2011} for sentiment
analysis), or modeling techniques designed for general use across
different settings and genres. Basic research tends to focus on one problem at a time, and frequently leverages established
datasets to provide a well-controlled environment for varying model design. Basic NLP research is intended to take the long view: it takes the time to investigate fundamental questions that may yield rewards for years to come.

\textbf{Applied research} Applied NLP research studies the intersection of
universal principles with specific settings; it is responsive to the needs
of commercial applications or researchers in other domains.
Applied research utilizes real-world datasets, often specialized, and involves sources of noise and unreliability that
complicate capturing linguistic regularities of interest. Applications often involve tackling multiple interrelated problems, and demand complex combinations of tools (e.g. using OCR followed by NLP to analyze scanned documents). Applied research is concrete and immediate, but may also be reactive and have a limited scope.

\textbf{Translational research} The term \textit{translational} is used in
medicine to describe research that aims to transform advances in basic knowledge
(biological or clinical) to applications to human health
\cite{Butte2008,Rubio2010}. Translational research is a distinct discipline
bridging basic science and applications \cite{Pober2001,Reis2010}. We adopt the
term \textit{Translational NLP} to describe research bridging the gap between 
basic and applied NLP research, and aiming to understand the processes by which
each informs the other.  Section~\ref{sec:case-studies} presents one in-depth
example; other salient examples include comparing the efficacy of domain adaptation
methods for different application domains \cite{Naik2019} and developing reusable
software for processing specific text genres \cite{Neumann2019}. Translational research occupies a middle ground in the timeframe and complexity of solutions: it develops processes to rapidly and effectively integrate new innovations into applications to address emerging needs, and facilitates integration between pipelines of NLP tools.

\subsection{Translation is bidirectional}
In addition to ``forward'' motion of basic innovations into practical applications, the needs of real-world applications also provide significant opportunities for new fundamental research.  Shneiderman's model of ``two parents, three children'' \cite{Shneiderman2016} provides an informative picture: combining a practical problem and a theoretical model yields (1) a solution to the problem, (2) a refinement of the theory, and (3) guidance for future research.
Tight links between basic research and applications have driven many major advances in NLP, from machine translation and dialog systems to search engines and question answering. Designing research with application needs in mind is a key impact criterion for both funding agencies \cite{Christianson2018} and industry \cite{Spector2012}, and helps to identify new, high-impact research problems \cite{Shneiderman2018}.

\subsection{NLP as a translational field: a historical perspective}

The NLP field has always lain at the nexus of basic and applied
research.  Application needs have driven some of the most fundamental
developments in the field, leading to explosions in basic research in new
topics and on long-standing challenges.

The need to automatically translate Russian scientific papers in the early years
of the Cold War led to some of the earliest NLP research, creating the
still-thriving field of machine translation \cite{Slocum1985}.  Machine translation
has since helped drive many significant advances in basic NLP
research, from the adoption of statistical models in the 1980s \cite{Dorr1999}
to neural sequence-to-sequence modeling \cite{Sutskever2014} and attention
mechanisms \cite{Bahdanau2015}.

Similarly, the rapid growth of the World Wide Web in the 1990s created an acute
need for technologies to search the growing sea of information, leading to the
development of NLP-based search engines such as Lycos \cite{Mauldin1997},
followed by PageRank \cite{Page1999} and the growth of Google. The need to index
and monetize vast quantities of textual information led to an explosion in
information retrieval research, and the NLP field and ever-growing web data
continue to co-develop.

In a more recent example, IBM identified automated question answering (QA) as a
new business opportunity in a high-information world, and developed the Watson
project \cite{Ferrucci2010}. Watson's early successes catapulted QA into the center of
NLP research, where it has continued to drive both novel technology development
and benchmark evaluation datasets used in hundreds of basic NLP studies
\cite{Rajpurkar2016}.

These and other examples illustrate the key role that application needs have
played in driving innovation in NLP research.  This reflects not only the
history of the field but the role that integrating basic and applied
research has in enriching scientific endeavor
\cite{Stokes1997,Branscomb1999,Narayanamurti2013,Shneiderman2016}.
An integrated approach has been cited by both Google \cite{Spector2012} and IBM
\cite{McQueeney2003} as central to their successes in both business and
research.  The aim of our paper is to facilitate this integration in NLP more
broadly, through presenting a rubric for studying and facilitating the process of getting both
to and back from application.

\subsection{A practical definition}
For an operational definition of Translational NLP, it is instructive to consider four phases of a generic workflow for tackling a novel NLP problem using supervised
machine learning.\footnote{
    While workflows will vary for different classes of NLP problems,
    dialogue between NLP experts and subject matter experts is at the
    heart of developing almost all NLP solutions.
}  First, a team of NLP experts
works with subject matter experts (SMEs) to identify appropriate corpora, define concepts
to be extracted, and construct annotation guidelines for the target task.  Second,
SMEs use these guidelines to annotate natural language data, using
iterative evaluation, revision of guidelines, and re-annotation to converge on a
high-quality \textit{gold standard} set of annotations.  Third, NLP experts use
these annotations to train and evaluate candidate models of the task, joined with
SMEs in a feedback loop to discuss results and needed revisions of
goals, guidelines, and gold standards.  Finally, buy-in is sought from SMEs and
practitioners in the target domain, in a dialogue informed by empirical results
and conceptual training.  NLP adoption in practice identifies failure cases and
new information needs, and the process begins again.

This laborious process is needed because of the gaps between expertise in NLP technology and expertise in use cases where NLP is applied.  NLP expertise is needed to properly formulate problems, and subsequently to develop sound and generalizable solutions to those problems. However, for uptake (and therefore impact) to occur, these solutions must be based in deep expertise in the use case domain, reified in a computable manner through annotation or knowledge resource development.  These distinct forms of expertise are generally found in different groups of individuals with complementary perspectives (see e.g. \citet{kruschwitz2017searching}).

Given this gap, we define \textbf{Translational NLP} as the development of theories, tools, and processes to enable the direct application of advanced NLP tools in specific use cases.  Implementing these tools and processes, and engaging with basic NLP experts and SMEs in their use, is the role of the Translational NLP scientist. Although every use case has unique characteristics, there are shared principles in designing NLP solutions that undergird the whole of the research and application process.  These shared translational principles can be adopted by basic researchers to increase the impact of NLP methods innovations, and guide the translational researcher in developing novel efforts targeting fundamental gaps between basic research and applications.  The framework presented in this paper identifies common variables and asks specific questions that can drive this research.

\begin{figure*}
    \centering
    \includegraphics[width=0.95\textwidth]{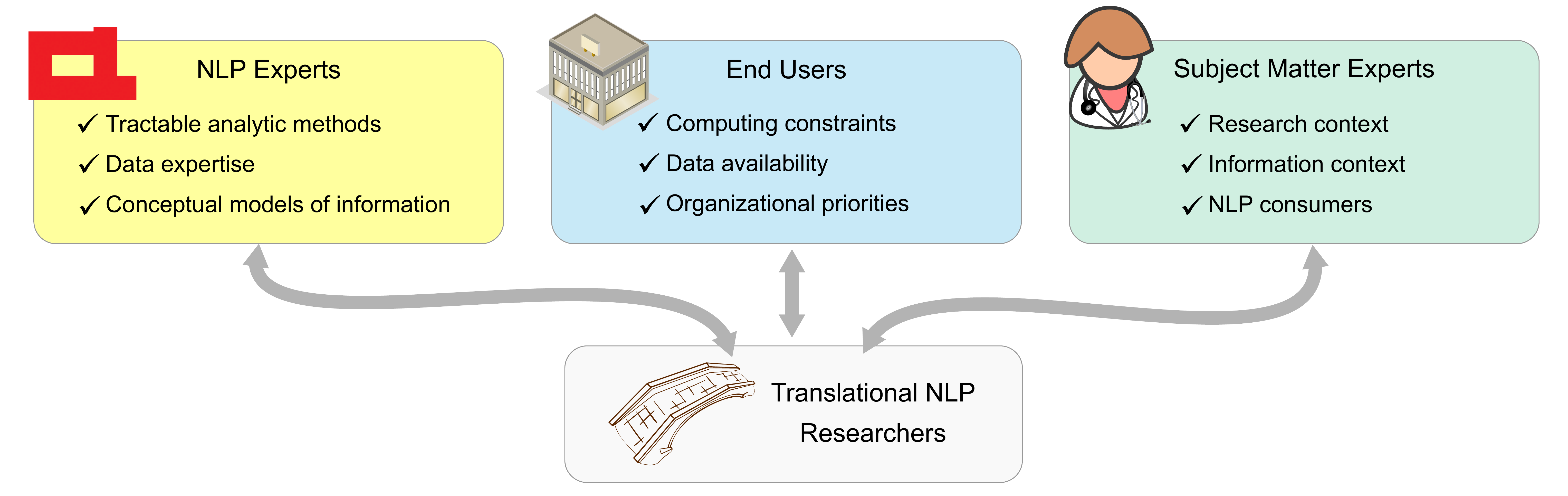}
    \caption{Attributes of key stakeholders in the translational process for NLP.}
    \label{fig:stakeholders}
\end{figure*}

For examples of this process in practice, it is valuable to examine NLP development in the medical domain.  Use-inspired NLP research has a long history in medicine \cite{Sager1982,Ranum1989}, frequently with an eye towards practical applications in research and care.  \citet{Chapman2011} highlight shared tasks as a key step towards addressing numerous barriers to application of NLP on clinical notes, including lack of shared datasets, insufficient conventions and standards, limited reproducibility, and lack of user-centered design (all factors presenting basic research opportunities, in addition to NLP task improvement). Several efforts have explored the development of graphical user interfaces for conducting NLP tasks, including creation and execution of pipelines~\cite{cunningham2002gate,davolio2010evaluation,davolio2011automated,Soysal2018}, although these efforts generally do not report on evaluation of usability by non-NLP experts. Usability has been investigated by other studies involving more focused tools aimed at specific NLP tasks, including concept searching~\cite{Hultman2018}, annotation~\cite{gobbel2014development,gobbel2014assisted}, and interactive review of and update of text classification models~\cite{Trivedi2018,Trivedi2019,savelka2015applying}.  Recent research has utilized interactive NLP tools for processing cancer research \cite{Deng2019} and care \cite{Yala2017} documents. By constructing, designing, and evaluating tools designed to simplify specific NLP processes, these efforts present examples of Translational NLP.

\section{The Translational NLP framework}
\label{sec:translational-nlp}

We present a conceptual framework for Translational NLP, to formalize shared principles describing how basic and applied research interact to create NLP solutions.
Our framework codifies fundamental variables in this process, providing a roadmap for negotiating the design of methodological innovations
with an eye towards potential applications.  Although it is certainly not the case that
every basic research advance must be tied to a downstream application need,
designing foundational technologies for potential application from the
beginning produces more robust technologies that are easier
to transfer to practical settings, increasing the impact of basic
research.  By defining common variables, our framework also provides a structure for aligning application needs to basic technologies, helping to
identify potential failure points and new research needs early for faster
adoption of basic NLP advances.

Our framework has two components:
\begin{enumerate}
    \item A definition of broad classes of \ul{stakeholders} in translating basic NLP innovations into applications, including the roles that each stakeholder plays
    in defining and guiding research;
    \item A \ul{checklist} of fundamental questions to structure the Translational NLP process,
    and to guide identification of
    basic research opportunities in specific application cases.
\end{enumerate}

\subsection{Stakeholders}
\label{sec:stakeholders}

NLP applications involve three broad categories of stakeholders, illustrated in Figure~\ref{fig:stakeholders}. Each contributes differently to technology implementation and identifying new
research challenges.

\textbf{NLP Experts} NLP researchers bring key \ul{analytic} skills to enable achieving the goals of an applied system. NLP experts provide
\textit{methodological sophistication} in models and paradigms for analyzing
language, and an understanding of the nature of language and how it captures
information. NLP researchers provide much-needed \textit{data expertise},
including skills in obtaining, cleaning, and formatting data for machine
learning and evaluation, as well as \textit{conceptual models for representing
information needs}.
NLP scientists identify research opportunities in modeling information
needs, bringing linguistic knowledge into the equation, and
developing appropriate tools for application and reuse.

\textbf{Subject Matter Experts} Subject matter experts (SMEs) provide the \ul{context} that helps to
determine what information is important to analyze and what the outputs of
applied NLP systems mean for the application setting.
SMEs, from medical practitioners to legal scholars and financial
experts, bring an understanding of \textit{where relevant information can be found}
(e.g., document sources \cite{Fisher2016a} and sections \cite{Afzal2018}), which
can help identify new types of language for basic researchers to study
\cite{Burstein2009,Crossley2014} and new challenges such as sparse complex
information \cite{Newman-Griffis2019emnlp} and higher-level structure in complex
documents \cite{Naik2019}.
In addition, the context that domain experts offer in terms of the needs of
target applications feeds back into \textit{evaluation} methods in the basic research
setting \cite{Graham2015}. 

SMEs are also the consumers of NLP solutions, as tools for their own research and applications.  Thus, SMEs must also be consultants regarding the trustworthiness and reliability of proposed solutions, and can identify key application-specific concerns such as security requirements.

\textbf{End Users} 
The end users of NLP solutions span a range of roles, environmental contexts, and goals, each of which guides \ul{implementation} factors of NLP applications. For example, collecting patient language in a lab setting, in a clinic, or at home will pose different challenges in each setting, which can inform the development of basic NLP methods.
Application settings may have limited
\textit{computational resources}, motivating the development of efficient
alternatives to high-resource models (e.g. \citet{Wang2020}), and have different {\it human factors} affecting information collection and use.

End users have different constraints on \textit{data
availability}, in terms of how much data of what types can be obtained from
whom; the extensive work funded by DARPA's Low Resource Languages for Emergent
Incidents (LORELEI) initiative \cite{Christianson2018} is a testament to the
basic research arising from these constraints.

Beyond the individual domain expert, end users use NLP technologies to address their own information needs according to the {\it priorities} of their organizations.  These organizational priorities may
conflict with existing modeling assumptions, highlighting new opportunities for
basic research to expand model capabilities.  For example, \citet{Shah2019}
highlight the conceptual gap between predictive model performance in medicine
and clinical utility to call for new research on utility-driven model
evaluation.  \citet{Spector2012} make a similar point about Google's
mission-driven research identifying unseen gaps for new basic research.

The role of the \textbf{Translational NLP} researcher is to interface with each
of these stakeholders, to connect their goals, constraints, and contributions
into a single applied system, and to identify new research opportunities where
parts of this system conflict with one another. Notably, this creates an opportunity for valuable study of SME and end user research practices, and for participatory design of NLP research \cite{lazar2017research}.  Our checklist, introduced in
the next section, provides a structured framework for this translational process.

\begin{figure}
    \centering
    \includegraphics[width=0.5\textwidth]{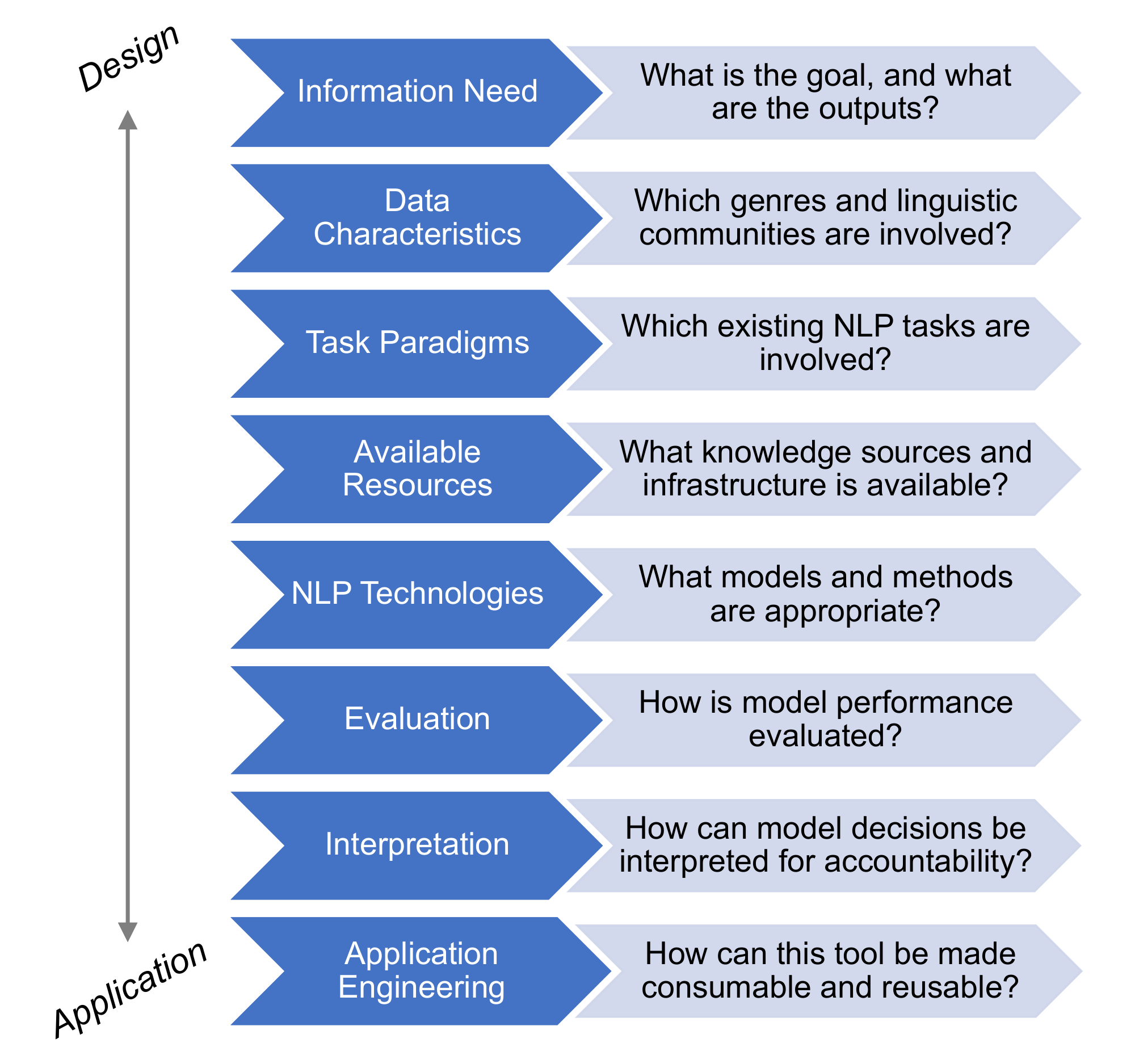}
    \caption{The eight items in our Translational NLP checklist, with key
    questions for each. Items are loosely ordered from initial design to
    application details, but should be regularly revisited in a feedback loop
    between application stakeholders.}
    \label{fig:checklist}
\end{figure}

\subsection{Translational NLP Checklist}
\label{sec:checklist}

The path between basic research and applications is often nebulous in NLP, limiting
the downstream impact of modeling innovations and obscuring basic research
challenges found in application settings.  We present a general-purpose checklist covering fundamental variables in translating basic research into applications,
which breaks down the translational process into discrete pieces for negotiation,
measurement, and identification of new research opportunities.
Our checklist, illustrated in
Figure~\ref{fig:checklist}, is loosely ordered from initial design to
application details. In practice, these items reflect different elements of the
application process and are constantly re-evaluated via a
feedback loop between the application stakeholders. While many of these items will
be familiar to NLP researchers, each represents potential points of failure in translation.
Designing the research process with these variables in mind will produce basic
innovations that are more easily adopted for application and more directly connected
to the challenges of real-world use cases.

We illustrate our items for two example cases:

\noindent {\it Ex.\ 1:} Analysis of multimodal clinical data (scanned text, tables, images) for patient diagnosis.

{\setlength{\parskip}{0pt}
\noindent {\it Ex.\ 2:} Comparison of medical observations to government treatment and billing guidelines.
}

\textbf{Information Need} The initial step that guides an application is
defining inputs and outputs, at two levels: (1) the \textit{overall problem}
to address with NLP (led by the subject matter expert), and (2) the \textit{formal
representation} of that problem (led by the NLP expert). The overall
goal (e.g., ``extract information on cancer from clinical notes'') determines the
requirements of the solution, and is central to identifying a measurement of its
effectiveness.  Once the overall goal is determined, the next step is a formal
representation of that goal in terms of text units (documents, spans) to
analyze and what the analysis should produce (class labels, sequence
annotations, document rankings, etc.). These requirements are tailored to
specific applications and may not reflect standardized NLP tasks. For example, a
clinician interested in the documented reasoning behind a series of laboratory
test orders needs: (1) the orders themselves (text spans); (2) the temporal
sequence of the orders; and (3) a text span containing the justification for
each order.

\noindent {\it Ex.\ 1:} type, severity, history of symptoms.

{\setlength{\parskip}{0pt}
\noindent {\it Ex.\ 2:} clinical findings, logical criteria.
}

\textbf{Data Characteristics} A clear description of the \textit{language data to be analyzed} is
key to identifying appropriate NLP technologies. Data characteristics include
the natural language(s) used (e.g., English, Chinese), the genre(s) of
language to analyze (e.g., scientific abstracts, quarterly earnings reports,
tweets, conversations), and the type(s) of linguistic community that produced them (e.g.,
medical practitioners, educators, policy experts).  This
information identifies the sublanguage(s) of interest \cite{Grishman1986}, which
determine the availability and development of appropriate NLP tools
\cite{Grishman2001}.  Corporate disclosures, financial news reports, and tweets all
require different processing strategies \cite{Xing2018}, as do tweets written
by different communities \cite{Blodgett2016,Groenwald2020}.

\noindent {\it Ex.\ 1:} clinical texts, lab reports.

\noindent {\it Ex.\ 2:} clinical texts, legal guidelines.

\textbf{Task Paradigms} To address the overall goal with an NLP solution, it must be formulated in terms of one or more \textit{well-defined NLP problems}.  Many real-world application needs do not clearly
correspond to a single benchmark task formulation.  For example, our earlier
example of the sequence of lab order justifications can be formulated as a
sequence of: (1) Named Entity
Recognition (treating the order types as named entities in a medical knowledge
base); (2) time expression extraction and normalization; (3) event ordering;
and (4) evidence identification.  Breaking the application need into
well-studied subproblems at design time enables faster identification and
development of relevant NLP technologies, and highlights any portions of the
goal that do \textit{not} correspond with a known problem, requiring
novel basic research.

\noindent {\it Ex.\ 1:} document type classification, OCR, information extraction (IE), patient classification.

{\setlength{\parskip}{0pt}
\noindent {\it Ex.\ 2:} IE, natural language inference.
}

\textbf{Available Resources} The question of resources to support an NLP solution includes two distinct concerns: (1) {\it knowledge sources} available to represent salient aspects of the target task; and (2) {\it compute infrastructure} for NLP system execution and deployment. Knowledge sources may be symbolic, such as knowledge graphs or gazetteers, or representational, such as representative corpora or pretrained language models. For some applications, powerful knowledge sources may be available (such as the UMLS \cite{Bodenreider2004} for biomedical reasoning), while others are severely under-resourced (such as emerging geopolitical events, which may lack even relevant social media text). These resources in turn affect the kinds of technologies that are appropriate to use.

In terms of infrastructure, NLP technologies are deployed on a wide variety of
systems, from commercial data centers to mobile devices. Each setting
presents constraints of \textit{limited resources} and \textit{throughput
requirements} \cite{nityasya2020no}.  An application environment with a high maximum resource load but low
median availability is amenable to batch processing architectures or approaches
with high pretraining cost and low test-time cost.  Pretrained word representstions
\cite{Mikolov2013a,Pennington2014} and language models \cite{Peters2018,Devlin2019}
are one example of fundamental technologies that address such a need.  Throughput requirements, i.e., how much language input needs to be analyzed in
a fixed amount of time, often require
engineering optimization for specific environments \cite{Afshar2019}, but the need
for faster runtime computation has led to many advances in machine learning for NLP,
such as variational autoencoders \cite{Kingma2014} and the Transformer architecture
\cite{Vaswani2017}.

\noindent {\it Ex.\ 1:} UMLS, high GPU compute.

{\setlength{\parskip}{0pt}
\noindent {\it Ex.\ 2:} UMLS, guideline criteria, low compute.
}

\textbf{NLP Technologies}
The interaction between task paradigms, data characteristics, and available resources
helps to determine what types of \textit{implementations} are appropriate to the task.
Implementations can be further broken down into \textit{representation technologies},
for mathematically representing the language units to be analyzed; \textit{modeling
architectures}, for capturing regularities within that language; and \textit{optimization
strategies} (when using machine learning), for efficiently estimating model parameters
from data.  In low-resource settings, highly parameterized models such as BERT may
not be appropriate, while large-scale GPU server farms enable highly complex model
architectures.  
When the overall goal is factorized into multiple NLP tasks,
optimization often involves joint or multi-task learning \cite{Caruana1997}.

\noindent {\it Ex.\ 1:} large language models, dictionary matching, OCR, multi-task learning.

{\setlength{\parskip}{0pt}
\noindent {\it Ex.\ 2:} dictionary matching, small neural models.
}

\textbf{Evaluation}
Once a solution has been designed, it must be evaluated in terms of
both \textit{the specific NLP problem(s)} and \textit{the overall goal}
of the application.  Standardized NLP task formulations typically define benchmark metrics which can be used for evaluating the NLP components: F-1 and
AUC for information extraction, MRR and NDCG for information retrieval, etc.
The design of these metrics is
its own extensive area of research \cite{Jones1996,Hirschman1997,Graham2015},
and even established evaluation methods may be
constantly revised \cite{Grishman1995}.
Critically for the translational researcher, some metrics may be preferred over others (e.g., precision over recall), and standardized evaluation metrics
may not reflect the goals and needs of
applications \cite{Friedman1998}. Improvements on standardized evaluation metrics (such as increased AUC) may even obscure degradations in application-relevant performance
measures (such as decreased process efficiency).
Translational researchers thus have the opportunity to work with NLP experts and SMEs
to identify or develop metrics that capture both the effectiveness of the NLP system
and its contribution to the application's overall goal.

\noindent {\it Ex.\ 1:} F-1, patient outcomes.

{\setlength{\parskip}{0pt}
\noindent {\it Ex.\ 2:} F-1, billing rates.
}

\textbf{Interpretation}
Interpretability and analysis of NLP and other machine learning systems has been the
focus of extensive research in recent years \cite{Gilpin2018,Belinkov2019},
with debate over what constitutes an interpretation \cite{Rudin2019,wiegreffe-pinter-2019-attention} and
development of broad-coverage software packages for ease of use \cite{Nori2019}.
For the translational researcher, the first step is to engage with SMEs to determine
\textit{what constitutes an acceptable interpretation} of an NLP system's output in the
application domain (which may be subject to specific legal or ethical
requirements around accountability in decision-making processes).
This leads to an iterative
process, working with SMEs and NLP experts to identify appropriately
interpretable models, or to identify the need for new basic research on interpretability
within the target domain.

\noindent {\it Ex.\ 1:} Evidence identification, model audits.

{\setlength{\parskip}{0pt}
\noindent {\it Ex.\ 2:} Criteria visualization, model audits.
}

\textbf{Application Engineering}
Last but not least, the translational process must also be concerned with the
\textit{implementation} of NLP solutions, both in terms of the specific technologies
used and how they can fit in to broader information processing pipelines.
The development of general-purpose NLP architectures such as the Stanford CoreNLP
Toolkit \cite{Manning2014}, spaCy \cite{spaCy}, and AllenNLP
\cite{Gardner2018allennlp}, as well as more targeted architectures such as the
clinical NLP framework presented by \citet{Wen2019},
provide well-engineered frameworks for implementing new
technologies in a way that is easy for others to both adopt and adapt for use in
their own pipelines.  Standardized data exchange frameworks such as UIMA
\cite{ferrucci2004uima} and JSON make implementations more modular and easier
to wire together.  Leveraging tools and frameworks like these, together with
good software design principles, makes NLP tools both easier to apply downstream
and easier for other researchers to incorporate into their own work.

\noindent {\it Ex.\ 1:} Multiple interoperable technologies.

{\setlength{\parskip}{0pt}
\noindent {\it Ex.\ 2:} Single decision support tool.
}

\subsubsection{Translating methodology advances into existing applications}
While the checklist items can guide initial design
of a new NLP solution, they are equally applicable for incorporating new basic
NLP innovations into existing solutions.  Any new innovation can be reviewed in
terms of our checklist items to identify new requirements or constraints
(e.g., higher computational cost, more intuitive interpretability measures).
The translational researcher can then work with NLP experts, SMEs, and the
end users to determine how to incorporate the new innovation
into the existing solution.

\begin{table*}
    \centering
    \footnotesize
    \begin{tabular}{p{2.5cm}p{12cm}}
        \toprule
        Information Need&
                \textit{Overall goal:} Improve disability benefits review process by highlighting relevant information
                
                \textit{Formal representation:} Spans of evidence, with attributes for activity type and level of limitation
            \\
        \midrule
        Data Characteristics&
            Medical records and administrative forms from  USA, mostly English
            \\
        \midrule
        Task Paradigms&
            Information extraction (spans), information retrieval (documents), span classification (activity and limitations)
            \\
        \midrule
        Available Resources&
            Minimal knowledge sources for function and disability, no public corpora; US government computing systems; high throughput requirements (thousands of records/day)
            \\
        \midrule
        NLP Technologies&
            Low-latency, low-compute sequence models; rule-based systems
            \\
        \midrule
        Evaluation&
            Standard metrics (F-1, accuracy). Information retrieval metrics reported for use case prototypes.
            \\
        \midrule
        Interpretation&
            Interpretation needs primarily around human decision-making; NLP tools  highlight and organize information in context. No ML interpretability reported in published results.
            \\
        \midrule
        Application
        
        Engineering&
            Open-source implementations using standardized frameworks for preprocessing. No data exchange reported.
            \\
        \bottomrule
    \end{tabular}
    \caption{Translational NLP checklist items for Disability Review case study, including notes on published results.}
    \label{tbl:case-study}
\end{table*}

\section{Case Study: NLP for Disability Review}
\label{sec:case-studies}

We illustrate our Translational NLP framework using our
recent line of research on developing NLP tools to assist US Social Security Administration (SSA)
officials in reviewing applications for disability benefits \cite{Desmet2020}.
The goal of this effort was to help identify relevant pieces of medical evidence
for making decisions about disability benefits, analyzing vast quantities of medical records
collected during the review process.

The stakeholders in this setting included: NLP researchers (interested in
developing generalizable methods); subject matter experts in disability and rehabilitation; and SSA end users (limited computing,
large data but strictly controlled, overall priorities of efficiency and
accuracy).

The Translational NLP checklist for this setting is shown in
Table~\ref{tbl:case-study}. This combination of factors has led
to several translational studies, including:
\vspace{-\topsep}
\begin{itemize}
    \setlength{\parskip}{0pt}
    \setlength{\itemsep}{0pt plus 1pt}
    \item \citet{Newman-Griffis2018repl4nlp} developed a low-resource entity embedding method for domains with minimal knowledge sources (lack of Available Resources).
    \item \citet{Newman-Griffis2018bionlp} analyzed the data size and representativeness tradeoff for information extraction in domains lacking large corpora (Available Resources).
    \item \citet{Newman-Griffis2019emnlp} developed a flexible method for identifying sparse health information that is syntactically complex (challenging Data Characteristics).
    \item \citet{newman-griffis2021automated} compared the Task Paradigms of classification and candidate selection paradigms for medical coding in a new domain.
\end{itemize}
\vspace{-\topsep}

While these studies do not systematically explore Evaluation, Interpretation, or Application
Engineering, they illustrate how the characteristics of one application setting can lead to a line of Translational
NLP research with broader implications. 
Several further challenges of this application area remain unstudied: for example, representing
and modeling the complex timelines of persons with chronic health conditions and
intermittent health care and adapting NLP systems to highly variable
medical language from practitioners and patients around the US.
These present intriguing challenges for basic NLP research that can inform many other applications beyond this case study.

Of course, these studies are far from the only examples of Translational NLP research. Many studies tackle translational questions, from domain adaptation (shifts in Data Characteristics) and low-resource learning (limited Available Resources), and the growing NLP literature in domain-specific venues such as medical research, law, finance, and more involves all aspects of the translational process. Rather, this case study is simply one illustration of how an explicitly translational perspective in study design can identify and connect broad opportunities for contributions to NLP research.

\section{Discussion}
\label{sec:discussion}

Our paradigm of Translational NLP defines and gives structure to a
valuable area of research not explicitly represented in the ACL community.
We note that translational research is not meant to replace either basic
or applied research, nor do we intend to say that all basic NLP studies must be tied to
specific application needs. Rather we aim to highlight the value of studying
the \textit{processes}
of turning basic innovations into successful applications.
These processes, from scaling model computation to redesigning tools to meet changing application needs, can inform new research in model design, domain adaptation, etc., and can help us understand why some tools succeed in application while others fail.
In addition to helping more innovations successfully translate, the principles outlined in this
paper can be of use to basic and applied NLP researchers as well as translational ones, in identifying
common variables and concerns to connect new work to the broader community.

Translational research is equally at home in industry and academia, and already occurring in both.  While resource disparities between industrial and academic research increasingly push large-scale modeling efforts out of reach of academic teams, a translational lens can help to identify rich areas of knowledge-driven study that do not require exascale data or computing resources.  The general principles and interdisciplinary nature of translational research make it a natural fit for public knowledge-driven academic settings, while its applicability to commercial needs is highly relevant to industry.

Our framework provides a starting point for the translational process, which will evolve differently for every project. The specifics of different applications will expand our initial questions in different ways (e.g., ``Data Characteristics'' may involve multimodal data, or different language styles), and the dynamics of collaborations will shift answers over time (e.g., a change in evaluation criteria may motivate different model training approaches). Our checklist provides a minimal set of common questions, and can function as a touchstone for discussions throughout the research process, but it can and should be tailored to the nature of each project. Our framework is itself a preliminary characterization of Translational NLP research, and will evolve over time as the field continues to develop.

\section{Conclusion}
We have outlined a new model of NLP research, Translational NLP,
which aims to bridge the gap between basic and applied NLP research
with generalizable principles, tools, and processes.  We identified
key types of stakeholders in NLP applications and how they inform the
translational process, and presented a checklist of common variables
and translational principles to consider in basic, translational, or
applied NLP research.  The translational framework reflects the central
role that integrating basic and applied research has played in the
development of the NLP field, and is illustrated by both the broad
successes of machine translation, speech processing, and web search,
as well as many individual studies in the ACL community and beyond.

\section*{Acknowledgments}
This work was supported by the National Library of Medicine of the National Institutes of Health under award number T15 LM007059, and National Science Foundation grant 1822831.

\typeout{}
\bibliography{references}

\begin{thebibliography}{88}
\expandafter\ifx\csname natexlab\endcsname\relax\def\natexlab#1{#1}\fi

\bibitem[{Afshar et~al.(2019)Afshar, Dligach, Sharma, Cai, Boyda, Birch,
  Valdez, Zelisko, Joyce, Modave, and Price}]{Afshar2019}
Majid Afshar, Dmitriy Dligach, Brihat Sharma, Xiaoyuan Cai, Jason Boyda, Steven
  Birch, Daniel Valdez, Suzan Zelisko, Cara Joyce, François Modave, and Ron
  Price. 2019.
\newblock \href {https://doi.org/10.1093/jamia/ocz068} {{Development and
  application of a high throughput natural language processing architecture to
  convert all clinical documents in a clinical data warehouse into standardized
  medical vocabularies}}.
\newblock \emph{Journal of the American Medical Informatics Association},
  26(11):1364--1369.

\bibitem[{Afzal et~al.(2018)Afzal, Mallipeddi, Sohn, Liu, Chaudhry, Scott,
  Kullo, and Arruda-Olson}]{Afzal2018}
Naveed Afzal, Vishnu~Priya Mallipeddi, Sunghwan Sohn, Hongfang Liu, Rajeev
  Chaudhry, Christopher~G Scott, Iftikhar~J Kullo, and Adelaide~M Arruda-Olson.
  2018.
\newblock \href
  {https://doi.org/https://doi.org/10.1016/j.ijmedinf.2017.12.024} {{Natural
  language processing of clinical notes for identification of critical limb
  ischemia}}.
\newblock \emph{International Journal of Medical Informatics}, 111:83--89.

\bibitem[{Anzaldi et~al.(2017)Anzaldi, Davison, Boyd, Leff, and
  Kharrazi}]{Anzaldi2017}
Laura~J Anzaldi, Ashwini Davison, Cynthia~M Boyd, Bruce Leff, and Hadi
  Kharrazi. 2017.
\newblock Comparing clinician descriptions of frailty and geriatric syndromes
  using electronic health records: a retrospective cohort study.
\newblock \emph{BMC Geriatrics}, 17(1):248.

\bibitem[{Bahdanau et~al.(2015)Bahdanau, Cho, and Bengio}]{Bahdanau2015}
Dzmitry Bahdanau, Kyunghyun Cho, and Yoshua Bengio. 2015.
\newblock \href {https://doi.org/10.1146/annurev.neuro.26.041002.131047}
  {{Neural Machine Translation By Jointly Learning To Align and Translate}}.
\newblock In \emph{ICLR 2015}.

\bibitem[{Belinkov and Glass(2019)}]{Belinkov2019}
Yonatan Belinkov and James Glass. 2019.
\newblock \href {https://doi.org/10.1162/tacl\_a\_00254} {Analysis methods in
  neural language processing: A survey}.
\newblock \emph{Transactions of the Association for Computational Linguistics},
  7:49--72.

\bibitem[{Blodgett et~al.(2016)Blodgett, Green, and O'Connor}]{Blodgett2016}
Su~Lin Blodgett, Lisa Green, and Brendan O'Connor. 2016.
\newblock \href {https://doi.org/10.18653/v1/D16-1120} {{Demographic Dialectal
  Variation in Social Media: A Case Study of African-American English}}.
\newblock In \emph{Proceedings of the 2016 Conference on Empirical Methods in
  Natural Language Processing}, pages 1119--1130, Austin, Texas. Association
  for Computational Linguistics.

\bibitem[{Bodenreider(2004)}]{Bodenreider2004}
Olivier Bodenreider. 2004.
\newblock \href {https://doi.org/10.1093/nar/gkh061} {{The Unified Medical
  Language System (UMLS): integrating biomedical terminology}}.
\newblock \emph{Nucleic Acids Research}, 32(90001):D267--D270.

\bibitem[{Branscomb(1999)}]{Branscomb1999}
Lewis~M Branscomb. 1999.
\newblock {The false dichotomy: Scientific creativity and utility}.
\newblock \emph{Issues in Science and Technology}, 16(1):66--72.

\bibitem[{Burstein(2009)}]{Burstein2009}
Jill Burstein. 2009.
\newblock {Opportunities for Natural Language Processing Research in
  Education}.
\newblock In \emph{Computational Linguistics and Intelligent Text Processing.
  CICLing 2009}, pages 6--27, Berlin, Heidelberg. Springer Berlin Heidelberg.

\bibitem[{Bush(1945)}]{Bush1945}
Vannevar Bush. 1945.
\newblock \emph{{Science, The Endless Frontier.}}
\newblock U.S. Government Printing Office.

\bibitem[{Butte(2008)}]{Butte2008}
Atul~J Butte. 2008.
\newblock \href {https://doi.org/10.1197/jamia.M2824} {{Translational
  Bioinformatics: Coming of Age}}.
\newblock \emph{Journal of the American Medical Informatics Association},
  15(6):709--714.

\bibitem[{Caruana(1997)}]{Caruana1997}
Rich Caruana. 1997.
\newblock Multitask learning.
\newblock \emph{Machine learning}, 28(1):41--75.

\bibitem[{Chapman et~al.(2011)Chapman, Nadkarni, Hirschman, D'Avolio, Savova,
  and Uzuner}]{Chapman2011}
Wendy~W Chapman, Prakash~M Nadkarni, Lynette Hirschman, Leonard~W D'Avolio,
  Guergana~K Savova, and Ozlem Uzuner. 2011.
\newblock \href {https://doi.org/10.1136/amiajnl-2011-000465} {{Overcoming
  barriers to NLP for clinical text: the role of shared tasks and the need for
  additional creative solutions}}.
\newblock \emph{Journal of the American Medical Informatics Association},
  18(5):540--543.

\bibitem[{Christianson et~al.(2018)Christianson, Duncan, and
  Onyshkevych}]{Christianson2018}
Caitlin Christianson, Jason Duncan, and Boyan Onyshkevych. 2018.
\newblock \href {https://doi.org/10.1007/s10590-017-9212-4} {{Overview of the
  DARPA LORELEI Program}}.
\newblock \emph{Machine Translation}, 32(1):3--9.

\bibitem[{Crossley et~al.(2014)Crossley, Allen, Kyle, and
  McNamara}]{Crossley2014}
Scott~A Crossley, Laura~K Allen, Kristopher Kyle, and Danielle~S McNamara.
  2014.
\newblock \href {https://doi.org/10.1080/0163853X.2014.910723} {{Analyzing
  Discourse Processing Using a Simple Natural Language Processing Tool}}.
\newblock \emph{Discourse Processes}, 51(5-6):511--534.

\bibitem[{Cunningham(2002)}]{cunningham2002gate}
Hamish Cunningham. 2002.
\newblock \href {https://doi.org/10.1023/A:1014348124664} {{GATE}, a {General}
  {Architecture} for {Text} {Engineering}}.
\newblock \emph{Computers and the Humanities}, 36(2):223--254.

\bibitem[{D'Avolio et~al.(2010)D'Avolio, Nguyen, Farwell, Chen, Fitzmeyer,
  Harris, and Fiore}]{davolio2010evaluation}
Leonard~W. D'Avolio, Thien~M. Nguyen, Wildon~R. Farwell, Yongming Chen, Felicia
  Fitzmeyer, Owen~M. Harris, and Louis~D. Fiore. 2010.
\newblock \href {https://doi.org/10.1136/jamia.2009.001412} {Evaluation of a
  generalizable approach to clinical information retrieval using the automated
  retrieval console ({ARC})}.
\newblock \emph{Journal of the American Medical Informatics Association},
  17(4):375--382.

\bibitem[{D'Avolio et~al.(2011)D'Avolio, Nguyen, Goryachev, and
  Fiore}]{davolio2011automated}
Leonard~W. D'Avolio, Thien~M. Nguyen, Sergey Goryachev, and Louis~D. Fiore.
  2011.
\newblock \href {https://doi.org/10.1136/amiajnl-2011-000183} {Automated
  concept-level information extraction to reduce the need for custom software
  and rules development}.
\newblock \emph{Journal of the American Medical Informatics Association},
  18(5):607--613.

\bibitem[{Deng et~al.(2019)Deng, Yin, Bao, Armengol, Wang, Tiwari, Barzilay,
  Parmigiani, Braun, and Hughes}]{Deng2019}
Zhengyi Deng, Kanhua Yin, Yujia Bao, Victor~Diego Armengol, Cathy Wang, Ankur
  Tiwari, Regina Barzilay, Giovanni Parmigiani, Danielle Braun, and Kevin~S.
  Hughes. 2019.
\newblock \href {https://doi.org/10.1200/CCI.19.00043} {Validation of a
  semiautomated natural language processing–based procedure for meta-analysis
  of cancer susceptibility gene penetrance}.
\newblock \emph{JCO Clinical Cancer Informatics}, (3):1--9.

\bibitem[{Desai et~al.(2020)Desai, Goh, Babu, and Aly}]{Desai2020}
Shrey Desai, Geoffrey Goh, Arun Babu, and Ahmed Aly. 2020.
\newblock Lightweight convolutional representations for on-device natural
  language processing.
\newblock \emph{arXiv preprint arXiv:2002.01535}.

\bibitem[{Desmet et~al.(2020)Desmet, Porcino, Zirikly, Newman-Griffis, Divita,
  and Rasch}]{Desmet2020}
Bart Desmet, Julia Porcino, Ayah Zirikly, Denis Newman-Griffis, Guy Divita, and
  Elizabeth Rasch. 2020.
\newblock \href {https://www.aclweb.org/anthology/2020.lt4gov-1.1} {Development
  of natural language processing tools to support determination of federal
  disability benefits in the {U}.{S}.}
\newblock In \emph{Proceedings of the 1st Workshop on Language Technologies for
  Government and Public Administration (LT4Gov)}, pages 1--6, Marseille,
  France. European Language Resources Association.

\bibitem[{Devlin et~al.(2019)Devlin, Chang, Lee, and Toutanova}]{Devlin2019}
Jacob Devlin, Ming-Wei Chang, Kenton Lee, and Kristina Toutanova. 2019.
\newblock \href {https://doi.org/10.18653/v1/N19-1423} {{BERT}: Pre-training of
  deep bidirectional transformers for language understanding}.
\newblock In \emph{Proceedings of the 2019 Conference of the North {A}merican
  Chapter of the Association for Computational Linguistics: Human Language
  Technologies, Volume 1 (Long and Short Papers)}, pages 4171--4186,
  Minneapolis, Minnesota. Association for Computational Linguistics.

\bibitem[{Dorr et~al.(1999)Dorr, Jordan, and Benoit}]{Dorr1999}
Bonnie~J Dorr, Pamela~W Jordan, and John~W Benoit. 1999.
\newblock \href {https://doi.org/https://doi.org/10.1016/S0065-2458(08)60282-X}
  {{A Survey of Current Paradigms in Machine Translation}}.
\newblock volume~49, pages 1--68. Elsevier.

\bibitem[{Elsahar and Gall{\'e}(2019)}]{Elsahar2019}
Hady Elsahar and Matthias Gall{\'e}. 2019.
\newblock \href {https://doi.org/10.18653/v1/D19-1222} {To annotate or not?
  predicting performance drop under domain shift}.
\newblock In \emph{Proceedings of the 2019 Conference on Empirical Methods in
  Natural Language Processing and the 9th International Joint Conference on
  Natural Language Processing (EMNLP-IJCNLP)}, pages 2163--2173, Hong Kong,
  China. Association for Computational Linguistics.

\bibitem[{Ferrucci et~al.(2010)Ferrucci, Brown, Chu-Carroll, Fan, Gondek,
  Kalyanpur, Lally, Murdock, Nyberg, Prager, and Others}]{Ferrucci2010}
David Ferrucci, Eric Brown, Jennifer Chu-Carroll, James Fan, David Gondek,
  Aditya~A Kalyanpur, Adam Lally, J~William Murdock, Eric Nyberg, John Prager,
  and Others. 2010.
\newblock {Building Watson: An overview of the DeepQA project}.
\newblock \emph{AI magazine}, 31(3):59--79.

\bibitem[{Ferrucci and Lally(2004)}]{ferrucci2004uima}
David Ferrucci and Adam Lally. 2004.
\newblock {UIMA}: an architectural approach to unstructured information
  processing in the corporate research environment.
\newblock \emph{Natural Language Engineering}, 10:1--26.

\bibitem[{Fisher et~al.(2016)Fisher, Garnsey, and Hughes}]{Fisher2016a}
Ingrid~E Fisher, Margaret~R Garnsey, and Mark~E Hughes. 2016.
\newblock \href {https://doi.org/https://doi.org/10.1002/isaf.1386} {{Natural
  Language Processing in Accounting, Auditing and Finance: A Synthesis of the
  Literature with a Roadmap for Future Research}}.
\newblock \emph{Intelligent Systems in Accounting, Finance and Management},
  23(3):157--214.

\bibitem[{Friedman and Hripcsak(1998)}]{Friedman1998}
C~Friedman and G~Hripcsak. 1998.
\newblock Evaluating natural language processors in the clinical domain.
\newblock \emph{Methods of information in medicine}, 37(4-5):334.

\bibitem[{Gardner et~al.(2018)Gardner, Grus, Neumann, Tafjord, Dasigi, Liu,
  Peters, Schmitz, and Zettlemoyer}]{Gardner2018allennlp}
Matt Gardner, Joel Grus, Mark Neumann, Oyvind Tafjord, Pradeep Dasigi,
  Nelson~F. Liu, Matthew Peters, Michael Schmitz, and Luke Zettlemoyer. 2018.
\newblock \href {https://doi.org/10.18653/v1/W18-2501} {{A}llen{NLP}: A deep
  semantic natural language processing platform}.
\newblock In \emph{Proceedings of Workshop for {NLP} Open Source Software
  ({NLP}-{OSS})}, pages 1--6, Melbourne, Australia. Association for
  Computational Linguistics.

\bibitem[{{Gilpin} et~al.(2018){Gilpin}, {Bau}, {Yuan}, {Bajwa}, {Specter}, and
  {Kagal}}]{Gilpin2018}
L.~H. {Gilpin}, D.~{Bau}, B.~Z. {Yuan}, A.~{Bajwa}, M.~{Specter}, and
  L.~{Kagal}. 2018.
\newblock \href {https://doi.org/10.1109/DSAA.2018.00018} {Explaining
  explanations: An overview of interpretability of machine learning}.
\newblock In \emph{2018 IEEE 5th International Conference on Data Science and
  Advanced Analytics (DSAA)}, pages 80--89.

\bibitem[{Gobbel et~al.(2014{\natexlab{a}})Gobbel, Garvin, Reeves, Cronin,
  Heavirland, Williams, Weaver, Jayaramaraja, Giuse, Speroff, Brown, Xu, and
  Matheny}]{gobbel2014assisted}
Glenn~T. Gobbel, Jennifer Garvin, Ruth Reeves, Robert~M. Cronin, Julia
  Heavirland, Jenifer Williams, Allison Weaver, Shrimalini Jayaramaraja, Dario
  Giuse, Theodore Speroff, Steven~H. Brown, Hua Xu, and Michael~E. Matheny.
  2014{\natexlab{a}}.
\newblock \href {https://doi.org/10.1136/amiajnl-2013-002255} {Assisted
  annotation of medical free text using {RapTAT}}.
\newblock \emph{Journal of the American Medical Informatics Association},
  21(5):833--841.

\bibitem[{Gobbel et~al.(2014{\natexlab{b}})Gobbel, Reeves, Jayaramaraja, Giuse,
  Speroff, Brown, Elkin, and Matheny}]{gobbel2014development}
Glenn~T. Gobbel, Ruth Reeves, Shrimalini Jayaramaraja, Dario Giuse, Theodore
  Speroff, Steven~H. Brown, Peter~L. Elkin, and Michael~E. Matheny.
  2014{\natexlab{b}}.
\newblock \href {https://doi.org/10.1016/j.jbi.2013.11.008} {Development and
  evaluation of {RapTAT}: a machine learning system for concept mapping of
  phrases from medical narratives}.
\newblock \emph{Journal of Biomedical Informatics}, 48:54--65.

\bibitem[{Graham(2015)}]{Graham2015}
Yvette Graham. 2015.
\newblock \href {https://doi.org/10.18653/v1/D15-1013} {{Re-evaluating
  Automatic Summarization with BLEU and 192 Shades of ROUGE}}.
\newblock In \emph{Proceedings of the 2015 Conference on Empirical Methods in
  Natural Language Processing}, pages 128--137, Lisbon, Portugal. Association
  for Computational Linguistics.

\bibitem[{Grishman(2001)}]{Grishman2001}
Ralph Grishman. 2001.
\newblock \href {http://nlp.cs.nyu.edu/publication/papers/grishman-ijcai01.pdf}
  {{Adaptive information extraction and sublanguage analysis}}.
\newblock In \emph{Proceedings of the Workshop on Adaptive Text Extraction and
  Mining, Seventeenth International Joint Conference on Artificial Intelligence
  (IJCAI-2001)}, pages 1--4, Seattle, Washington, USA.

\bibitem[{Grishman and Kittredge(1986)}]{Grishman1986}
Ralph Grishman and R~Kittredge. 1986.
\newblock \emph{{Analyzing language in restricted domains: Sublanguage
  description and processing}}.
\newblock Lawrence Erlbaum Associates.

\bibitem[{Grishman and Sundheim(1995)}]{Grishman1995}
Ralph Grishman and Beth Sundheim. 1995.
\newblock {Design of the MUC-6 evaluation}.
\newblock In \emph{Proceedings of the 6th Conference on Message Understanding},
  pages 1--11. Association for Computational Linguistics.

\bibitem[{Groenwold et~al.(2020)Groenwold, Ou, Parekh, Honnavalli, Levy, Mirza,
  and Wang}]{Groenwald2020}
Sophie Groenwold, Lily Ou, Aesha Parekh, Samhita Honnavalli, Sharon Levy, Diba
  Mirza, and William~Yang Wang. 2020.
\newblock \href {https://www.aclweb.org/anthology/2020.emnlp-main.473}
  {{Investigating African-American Vernacular English in Transformer-Based Text
  Generation}}.
\newblock In \emph{Proceedings of the 2020 Conference on Empirical Methods in
  Natural Language Processing (EMNLP)}, pages 5877--5883, Online. Association
  for Computational Linguistics.

\bibitem[{Hirschman and Thompson(1997)}]{Hirschman1997}
Lynette Hirschman and Henry~S. Thompson. 1997.
\newblock Overview of evaluation in speech and natural language processing.
\newblock In Ronald~A. Cole, editor, \emph{Survey of the State of the Art in
  Human Language Technology}. Cambridge University Press.

\bibitem[{Honnibal and Montani(2017)}]{spaCy}
Matthew Honnibal and Ines Montani. 2017.
\newblock spacy 2: Natural language understanding with bloom embeddings,
  convolutional neural networks and incremental parsing.

\bibitem[{Hultman et~al.(2018)Hultman, McEwan, Pakhomov, Lindemann, Skube, and
  Melton}]{Hultman2018}
Gretchen Hultman, Reed McEwan, Serguei Pakhomov, Elizabeth Lindemann, Steven
  Skube, and Genevieve~B. Melton. 2018.
\newblock Usability {Evaluation} of an {Unstructured} {Clinical} {Document}
  {Query} {Tool} for {Researchers}.
\newblock \emph{AMIA Joint Summits on Translational Science proceedings. AMIA
  Joint Summits on Translational Science}, 2017:84--93.

\bibitem[{Jones and Galliers(1996)}]{Jones1996}
Karen~Sp\"{a}rck Jones and Julia~R. Galliers. 1996.
\newblock \emph{Evaluating Natural Language Processing Systems: An Analysis and
  Review}.
\newblock Springer-Verlag, Berlin, Heidelberg.

\bibitem[{Kingma and Welling(2014)}]{Kingma2014}
Diederik~P Kingma and Max Welling. 2014.
\newblock Auto-encoding variational bayes.
\newblock \emph{ICLR 2014}.

\bibitem[{Kouloumpis et~al.(2011)Kouloumpis, Wilson, and Moore}]{Koulompis2011}
Efthymios Kouloumpis, Theresa Wilson, and Johanna Moore. 2011.
\newblock Twitter sentiment analysis: The good the bad and the omg!
\newblock In \emph{Fifth International AAAI conference on weblogs and social
  media}. Citeseer.

\bibitem[{Kruschwitz and Hull(2017)}]{kruschwitz2017searching}
Udo Kruschwitz and Charlie Hull. 2017.
\newblock \href {https://doi.org/10.1561/1500000053} {{Searching the
  Enterprise}}.
\newblock \emph{Foundations and Trends in Information Retrieval}, 11(1):18.

\bibitem[{Lazar et~al.(2017)Lazar, Feng, and Hochheiser}]{lazar2017research}
Jonathan Lazar, Jinjuan~Heidi Feng, and Harry Hochheiser. 2017.
\newblock \emph{{Research methods in human-computer interaction}}.
\newblock Morgan Kaufmann.

\bibitem[{Manning et~al.(2014)Manning, Surdeanu, Bauer, Finkel, Bethard, and
  McClosky}]{Manning2014}
Christopher Manning, Mihai Surdeanu, John Bauer, Jenny Finkel, Steven Bethard,
  and David McClosky. 2014.
\newblock \href {https://doi.org/10.3115/v1/P14-5010} {The {S}tanford
  {C}ore{NLP} natural language processing toolkit}.
\newblock In \emph{Proceedings of 52nd Annual Meeting of the Association for
  Computational Linguistics: System Demonstrations}, pages 55--60, Baltimore,
  Maryland. Association for Computational Linguistics.

\bibitem[{Mauldin(1997)}]{Mauldin1997}
M~I Mauldin. 1997.
\newblock \href {https://doi.org/10.1109/64.577466} {{Lycos: design choices in
  an Internet search service}}.
\newblock \emph{IEEE Expert}, 12(1):8--11.

\bibitem[{McQueeney(2003)}]{McQueeney2003}
David~F McQueeney. 2003.
\newblock {IBM's evolving research strategy}.
\newblock \emph{Research-Technology Management}, 46(4):20--27.

\bibitem[{Mikolov et~al.(2013)Mikolov, Chen, Corrado, and Dean}]{Mikolov2013a}
Tomas Mikolov, Kai Chen, Greg Corrado, and Jeffrey Dean. 2013.
\newblock \href {http://arxiv.org/abs/1301.3781} {{Efficient Estimation of Word
  Representations in Vector Space}}.
\newblock \emph{arXiv preprint arXiv:1301.3781}, pages 1--12.

\bibitem[{Naik et~al.(2019)Naik, Breitfeller, and Rose}]{Naik2019}
Aakanksha Naik, Luke Breitfeller, and Carolyn Rose. 2019.
\newblock \href {https://doi.org/10.18653/v1/W19-5929} {{TDDiscourse: A Dataset
  for Discourse-Level Temporal Ordering of Events}}.
\newblock In \emph{Proceedings of the 20th Annual SIGdial Meeting on Discourse
  and Dialogue}, pages 239--249, Stockholm, Sweden. Association for
  Computational Linguistics.

\bibitem[{Narayanamurti et~al.(2013)Narayanamurti, Odumosu, and
  Vinsel}]{Narayanamurti2013}
Venkatesh Narayanamurti, Tolu Odumosu, and Lee Vinsel. 2013.
\newblock {RIP: The basic/applied research dichotomy}.
\newblock \emph{Issues in Science and Technology}, 29(2):31--36.

\bibitem[{Neumann et~al.(2019)Neumann, King, Beltagy, and Ammar}]{Neumann2019}
Mark Neumann, Daniel King, Iz~Beltagy, and Waleed Ammar. 2019.
\newblock \href {https://doi.org/10.18653/v1/W19-5034} {{S}cispa{C}y: Fast and
  robust models for biomedical natural language processing}.
\newblock In \emph{Proceedings of the 18th BioNLP Workshop and Shared Task},
  pages 319--327, Florence, Italy. Association for Computational Linguistics.

\bibitem[{Newman-Griffis and Fosler-Lussier(2019)}]{Newman-Griffis2019emnlp}
Denis Newman-Griffis and Eric Fosler-Lussier. 2019.
\newblock \href {https://www.aclweb.org/anthology/D19-3015} {{HARE: a Flexible
  Highlighting Annotator for Ranking and Exploration}}.
\newblock In \emph{Proceedings of the 2019 Conference on Empirical Methods in
  Natural Language Processing and the 9th International Joint Conference on
  Natural Language Processing (EMNLP-IJCNLP): System Demonstrations}, pages
  85--90, Hong Kong, China. Association for Computational Linguistics.

\bibitem[{Newman-Griffis and
  Fosler-Lussier(2021)}]{newman-griffis2021automated}
Denis Newman-Griffis and Eric Fosler-Lussier. 2021.
\newblock \href {https://doi.org/10.3389/fdgth.2021.620828} {{Automated Coding
  of Under-Studied Medical Concept Domains: Linking Physical Activity Reports
  to the International Classification of Functioning, Disability, and Health}}.
\newblock \emph{Frontiers in Digital Health}, 3:620828.

\bibitem[{Newman-Griffis et~al.(2018)Newman-Griffis, Lai, and
  Fosler-Lussier}]{Newman-Griffis2018repl4nlp}
Denis Newman-Griffis, Albert~M Lai, and Eric Fosler-Lussier. 2018.
\newblock \href {https://doi.org/10.18653/v1/W18-3026} {Jointly embedding
  entities and text with distant supervision}.
\newblock In \emph{Proceedings of The Third Workshop on Representation Learning
  for {NLP}}, pages 195--206, Melbourne, Australia. Association for
  Computational Linguistics.

\bibitem[{Newman-Griffis and Zirikly(2018)}]{Newman-Griffis2018bionlp}
Denis Newman-Griffis and Ayah Zirikly. 2018.
\newblock \href {https://doi.org/10.18653/v1/W18-2301} {Embedding transfer for
  low-resource medical named entity recognition: A case study on patient
  mobility}.
\newblock In \emph{Proceedings of the {B}io{NLP} 2018 workshop}, pages 1--11,
  Melbourne, Australia. Association for Computational Linguistics.

\bibitem[{Nityasya et~al.(2020)Nityasya, Wibowo, Prasojo, and
  Aji}]{nityasya2020no}
Made~Nindyatama Nityasya, Haryo~Akbarianto Wibowo, Radityo~Eko Prasojo, and
  Alham~Fikri Aji. 2020.
\newblock {No Budget? Don't Flex! Cost Consideration when Planning to Adopt NLP
  for Your Business}.
\newblock \emph{arXiv preprint arXiv:2012.08958}.

\bibitem[{Nori et~al.(2019)Nori, Jenkins, Koch, and Caruana}]{Nori2019}
Harsha Nori, Samuel Jenkins, Paul Koch, and Rich Caruana. 2019.
\newblock {InterpretML: A unified framework for machine learning
  interpretability}.
\newblock \emph{arXiv preprint arXiv:1909.09223}.

\bibitem[{Page et~al.(1999)Page, Brin, Motwani, and Winograd}]{Page1999}
Lawrence Page, Sergey Brin, Rajeev Motwani, and Terry Winograd. 1999.
\newblock {The PageRank citation ranking: Bringing order to the web.}
\newblock Technical report, Stanford InfoLab.

\bibitem[{Pennington et~al.(2014)Pennington, Socher, and
  Manning}]{Pennington2014}
Jeffrey Pennington, Richard Socher, and Christopher Manning. 2014.
\newblock \href {https://doi.org/10.3115/v1/D14-1162} {{G}lo{V}e: Global
  vectors for word representation}.
\newblock In \emph{Proceedings of the 2014 Conference on Empirical Methods in
  Natural Language Processing ({EMNLP})}, pages 1532--1543, Doha, Qatar.
  Association for Computational Linguistics.

\bibitem[{Peters et~al.(2018)Peters, Neumann, Iyyer, Gardner, Clark, Lee, and
  Zettlemoyer}]{Peters2018}
Matthew Peters, Mark Neumann, Mohit Iyyer, Matt Gardner, Christopher Clark,
  Kenton Lee, and Luke Zettlemoyer. 2018.
\newblock \href {https://doi.org/10.18653/v1/N18-1202} {Deep contextualized
  word representations}.
\newblock In \emph{Proceedings of the 2018 Conference of the North {A}merican
  Chapter of the Association for Computational Linguistics: Human Language
  Technologies, Volume 1 (Long Papers)}, pages 2227--2237, New Orleans,
  Louisiana. Association for Computational Linguistics.

\bibitem[{Pober et~al.(2001)Pober, Neuhauser, and Pober}]{Pober2001}
Jordan~S Pober, Crystal~S Neuhauser, and Jeremy~M Pober. 2001.
\newblock {Obstacles facing translational research in academic medical
  centers}.
\newblock \emph{The FASEB Journal}, 15(13):2303--2313.

\bibitem[{Rahman et~al.(2020)Rahman, Nandi, and Hebert}]{Rahman2020}
Protiva Rahman, Arnab Nandi, and Courtney Hebert. 2020.
\newblock \href {https://doi.org/10.2196/19612} {Amplifying domain expertise in
  clinical data pipelines}.
\newblock \emph{JMIR Med Inform}, 8(11):e19612.

\bibitem[{Rajpurkar et~al.(2016)Rajpurkar, Zhang, Lopyrev, and
  Liang}]{Rajpurkar2016}
Pranav Rajpurkar, Jian Zhang, Konstantin Lopyrev, and Percy Liang. 2016.
\newblock {SQuAD: 100,000+ Questions for Machine Comprehension of Text}.
\newblock In \emph{Proceedings of the 2016 Conference on Empirical Methods in
  Natural Language Processing}, pages 2383--2392.

\bibitem[{Ranum(1989)}]{Ranum1989}
David~L Ranum. 1989.
\newblock Knowledge-based understanding of radiology text.
\newblock \emph{Computer methods and programs in biomedicine},
  30(2-3):209--215.

\bibitem[{Recasens and Hovy(2009)}]{Recasens2009}
Marta Recasens and Eduard Hovy. 2009.
\newblock \href {https://doi.org/10.1007/978-3-642-04975-0_3} {A deeper look
  into features for coreference resolution}.
\newblock In \emph{Proceedings of the 7th Discourse Anaphora and Anaphor
  Resolution Colloquium on Anaphora Processing and Applications}, DAARC '09,
  page 29–42, Berlin, Heidelberg. Springer-Verlag.

\bibitem[{Reis et~al.(2010)Reis, Berglund, Bernard, Califf, Fitzgerald,
  Johnson, Consortium, and Awards}]{Reis2010}
Steven~E Reis, Lars Berglund, Gordon~R Bernard, Robert~M Califf, Garret~A
  Fitzgerald, Peter~C Johnson, National~Clinical Consortium, and
  Translational~Science Awards. 2010.
\newblock \href {https://doi.org/10.1097/ACM.0b013e3181ccc877} {{Reengineering
  the national clinical and translational research enterprise: the strategic
  plan of the National Clinical and Translational Science Awards Consortium}}.
\newblock \emph{Academic Medicine}, 85(3):463--469.

\bibitem[{Rubio et~al.(2010)Rubio, Schoenbaum, Lee, Schteingart, Marantz,
  Anderson, Platt, Baez, and Esposito}]{Rubio2010}
Doris~McGartland Rubio, Ellie~E Schoenbaum, Linda~S Lee, David~E Schteingart,
  Paul~R Marantz, Karl~E Anderson, Lauren~Dewey Platt, Adriana Baez, and Karin
  Esposito. 2010.
\newblock \href {https://doi.org/10.1097/ACM.0b013e3181ccd618} {{Defining
  translational research: implications for training}}.
\newblock \emph{Academic Medicine}, 85(3):470--475.

\bibitem[{Rudin(2019)}]{Rudin2019}
Cynthia Rudin. 2019.
\newblock Stop explaining black box machine learning models for high stakes
  decisions and use interpretable models instead.
\newblock \emph{Nature Machine Intelligence}, 1(5):206--215.

\bibitem[{Sager et~al.(1982)Sager, Bross, Story, Bastedo, Marsh, and
  Shedd}]{Sager1982}
N~Sager, IDJ Bross, G~Story, P~Bastedo, E~Marsh, and D~Shedd. 1982.
\newblock Automatic encoding of clinical narrative.
\newblock \emph{Computers in Biology and Medicine}, 12(1):43--56.

\bibitem[{Savelka et~al.(2015)Savelka, Trivedi, and
  Ashley}]{savelka2015applying}
Jaromir Savelka, Gaurav Trivedi, and Kevin Ashley. 2015.
\newblock Applying an interactive machine learning approach to statutory
  analysis.
\newblock In \emph{JURIX 2015 - the 28th International Conference on Legal
  Knowledge and Information Systems}.

\bibitem[{Shah et~al.(2019)Shah, Milstein, and {Bagley Steven C.}}]{Shah2019}
Nigam~H Shah, Arnold Milstein, and PhD {Bagley Steven C.} 2019.
\newblock \href {https://doi.org/10.1001/jama.2019.10306} {{Making Machine
  Learning Models Clinically Useful}}.
\newblock \emph{JAMA}, 322(14):1351--1352.

\bibitem[{Shneiderman(2016)}]{Shneiderman2016}
Ben Shneiderman. 2016.
\newblock \emph{{The new ABCs of research: Achieving breakthrough
  collaborations}}.
\newblock Oxford University Press.

\bibitem[{Shneiderman(2018)}]{Shneiderman2018}
Ben Shneiderman. 2018.
\newblock \href {https://doi.org/10.1073/pnas.1802918115} {{Twin-Win Model: A
  human-centered approach to research success}}.
\newblock \emph{Proceedings of the National Academy of Sciences of the United
  States of America}, 115(50):12590--12594.

\bibitem[{Slater et~al.(2017)Slater, Joksimovi{\'c}, Kovanovic, Baker, and
  Gasevic}]{Slater2017}
Stefan Slater, Sre{\'c}ko Joksimovi{\'c}, Vitomir Kovanovic, Ryan~S Baker, and
  Dragan Gasevic. 2017.
\newblock Tools for educational data mining: A review.
\newblock \emph{Journal of Educational and Behavioral Statistics},
  42(1):85--106.

\bibitem[{Slocum(1985)}]{Slocum1985}
Jonathan Slocum. 1985.
\newblock {A Survey of Machine Translation: Its History, Current Status, and
  Future Prospects}.
\newblock \emph{Comput. Linguist.}, 11(1):1--17.

\bibitem[{Soysal et~al.(2018)Soysal, Wang, Jiang, Wu, Pakhomov, Liu, and
  Xu}]{Soysal2018}
Ergin Soysal, Jingqi Wang, Min Jiang, Yonghui Wu, Serguei Pakhomov, Hongfang
  Liu, and Hua Xu. 2018.
\newblock \href {https://doi.org/10.1093/jamia/ocx132} {{CLAMP} – a toolkit
  for efficiently building customized clinical natural language processing
  pipelines}.
\newblock \emph{Journal of the American Medical Informatics Association},
  25(3):331--336.
\newblock Publisher: Oxford Academic.

\bibitem[{Spector et~al.(2012)Spector, Norvig, and Petrov}]{Spector2012}
Alfred Spector, Peter Norvig, and Slav Petrov. 2012.
\newblock {Google's hybrid approach to research}.
\newblock \emph{Communications of the ACM}, 55(7):34--37.

\bibitem[{Stokes(1997)}]{Stokes1997}
Donald~E Stokes. 1997.
\newblock \emph{{Pasteur's quadrant: Basic science and technological
  innovation}}.
\newblock Brookings Institution Press.

\bibitem[{Sutskever et~al.(2014)Sutskever, Vinyals, and Le}]{Sutskever2014}
Ilya Sutskever, Oriol Vinyals, and Quoc~V Le. 2014.
\newblock {Sequence to sequence learning with neural networks}.
\newblock In \emph{Advances in neural information processing systems}, pages
  3104--3112.

\bibitem[{Trivedi et~al.(2019)Trivedi, Dadashzadeh, Handzel, Chapman,
  Visweswaran, and Hochheiser}]{Trivedi2019}
Gaurav Trivedi, Esmaeel~R Dadashzadeh, Robert~M Handzel, Wendy~W Chapman, Shyam
  Visweswaran, and Harry Hochheiser. 2019.
\newblock \href {https://doi.org/10.1055/s-0039-1695791} {{Interactive NLP in
  Clinical Care: Identifying Incidental Findings in Radiology Reports}}.
\newblock \emph{Appl Clin Inform}, 10(04):655--669.

\bibitem[{Trivedi et~al.(2018)Trivedi, Pham, Chapman, Hwa, Wiebe, and
  Hochheiser}]{Trivedi2018}
Gaurav Trivedi, Phuong Pham, Wendy~W. Chapman, Rebecca Hwa, Janyce Wiebe, and
  Harry Hochheiser. 2018.
\newblock \href {https://doi.org/10.1093/jamia/ocx070} {{NLPReViz: an
  interactive tool for natural language processing on clinical text}}.
\newblock \emph{Journal of the American Medical Informatics Association},
  25(1):81--87.

\bibitem[{Vaswani et~al.(2017)Vaswani, Shazeer, Parmar, Uszkoreit, Jones,
  Gomez, Kaiser, and Polosukhin}]{Vaswani2017}
Ashish Vaswani, Noam Shazeer, Niki Parmar, Jakob Uszkoreit, Llion Jones,
  Aidan~N Gomez, {\L}ukasz Kaiser, and Illia Polosukhin. 2017.
\newblock Attention is all you need.
\newblock In \emph{Advances in neural information processing systems}, pages
  5998--6008.

\bibitem[{Wang et~al.(2020)Wang, Wu, Liu, Cai, Zhu, Gan, and Han}]{Wang2020}
Hanrui Wang, Zhanghao Wu, Zhijian Liu, Han Cai, Ligeng Zhu, Chuang Gan, and
  Song Han. 2020.
\newblock {Hat: Hardware-aware transformers for efficient natural language
  processing}.
\newblock \emph{arXiv preprint arXiv:2005.14187}.

\bibitem[{Wen et~al.(2019)Wen, Fu, Moon, El~Wazir, Rosenbaum, Kaggal, Liu,
  Sohn, Liu, and Fan}]{Wen2019}
Andrew Wen, Sunyang Fu, Sungrim Moon, Mohamed El~Wazir, Andrew Rosenbaum,
  Vinod~C Kaggal, Sijia Liu, Sunghwan Sohn, Hongfang Liu, and Jungwei Fan.
  2019.
\newblock Desiderata for delivering nlp to accelerate healthcare ai advancement
  and a mayo clinic nlp-as-a-service implementation.
\newblock \emph{NPJ Digital Medicine}, 2(1):1--7.

\bibitem[{Wiegreffe and Pinter(2019)}]{wiegreffe-pinter-2019-attention}
Sarah Wiegreffe and Yuval Pinter. 2019.
\newblock \href {https://doi.org/10.18653/v1/D19-1002} {Attention is not not
  explanation}.
\newblock In \emph{Proceedings of the 2019 Conference on Empirical Methods in
  Natural Language Processing and the 9th International Joint Conference on
  Natural Language Processing (EMNLP-IJCNLP)}, pages 11--20, Hong Kong, China.
  Association for Computational Linguistics.

\bibitem[{Xing et~al.(2018)Xing, Cambria, and Welsch}]{Xing2018}
Frank~Z Xing, Erik Cambria, and Roy~E Welsch. 2018.
\newblock \href {https://doi.org/10.1007/s10462-017-9588-9} {{Natural language
  based financial forecasting: a survey}}.
\newblock \emph{Artificial Intelligence Review}, 50(1):49--73.

\bibitem[{Yala et~al.(2017)Yala, Barzilay, Salama, Griffin, Sollender, Bardia,
  Lehman, Buckley, Coopey, Polubriaginof et~al.}]{Yala2017}
Adam Yala, Regina Barzilay, Laura Salama, Molly Griffin, Grace Sollender,
  Aditya Bardia, Constance Lehman, Julliette~M Buckley, Suzanne~B Coopey,
  Fernanda Polubriaginof, et~al. 2017.
\newblock Using machine learning to parse breast pathology reports.
\newblock \emph{Breast Cancer Research and Treatment}, 161(2):203--211.

\end{thebibliography}
\bibliographystyle{acl_natbib}

\end{document}